%% file: paper.tex
\PassOptionsToPackage{bookmarks=false}{hyperref}
\documentclass[conference]{IEEEtran}

\usepackage[utf8]{inputenc}
\usepackage[T1]{fontenc}
\usepackage{xcolor}

\usepackage{microtype}
\usepackage{listings}
\definecolor{commentcolor}{rgb}{0.5,0.5,0.5}
\definecolor{darkgreen}{rgb}{0.09, 0.45, 0.27}
\definecolor{bgcolor}{rgb}{0.99,0.99,0.99}
\lstdefinestyle{adaptation}{
    basicstyle=\sffamily\footnotesize,
    keywords={if,then,else,estimator,return},
}
\lstdefinestyle{specification}{
    basicstyle=\sffamily\footnotesize,
    keywords={component,field,input,output,guard,condition,estimator,if,then,regression,time,to,type,ensemble,static,role,dynamic,with,cardinality,membership,T,in,unlimited,action,utility,priority},
}
\newcommand{\longlstinline}[1]{{\small\textsf{#1}}}
\newcommand{\footnotelstinline}[1]{{\scriptsize\textsf{#1}}}

\usepackage{paralist}
\usepackage{multicol}
\usepackage[inline]{enumitem}
\usepackage[normalem]{ulem}
\usepackage{url}
\usepackage{graphicx}
\usepackage{cite}

\makeatletter
\let\MYcaption\@makecaption
\makeatother
\usepackage[font=footnotesize]{subcaption}
\makeatletter
\let\@makecaption\MYcaption
\makeatother

\let\sss\subsubsection
\renewcommand{\subsubsection}[1]{\vspace{2mm}\sss{#1}}

\usepackage{fancyhdr}
\fancypagestyle{plain}{%
\fancyhead{}
\fancyfoot{}

\fancyfoot[L]{\scriptsize This is the authors' version of the paper: \textit{M. Töpfer, F. Plášil, T. Bureš, P. Hnětynka, M. Kruliš, D. Weyns:
Online ML Self-adaptation in Face of Traps, accepted for publication in Proceedings of ACSOS 2023, Toronto, Canada}.}
}
\fancypagestyle{empty}{%
\fancyhead{}
\fancyfoot{}

\fancyfoot[L]{\scriptsize This is the authors' version of the paper: \textit{M. Töpfer, F. Plášil, T. Bureš, P. Hnětynka, M. Kruliš, D. Weyns:
Online ML Self-adaptation in Face of Traps, accepted for publication in Proceedings of ACSOS 2023, Toronto, Canada}.}
}

\begin{document}

\clearpage
\pagestyle{plain}

\bstctlcite{IEEEexample:BSTcontrol}
\title{Online ML Self-adaptation in Face of Traps}

\author{%
	\IEEEauthorblockN{%
	    Michal Töpfer\IEEEauthorrefmark{1},
        František Plášil\IEEEauthorrefmark{1},
        Tomáš Bureš\IEEEauthorrefmark{1},
        Petr Hnětynka\IEEEauthorrefmark{1},
        Martin Kruliš\IEEEauthorrefmark{1},
        Danny Weyns\IEEEauthorrefmark{2}
	}%
	\IEEEauthorblockA{%
        \IEEEauthorrefmark{1}\textit{Charles University, Faculty of Mathematics and Physics, Prague, Czech Republic}\\
        \IEEEauthorrefmark{2}\textit{Katholieke Universiteit Leuven, Belgium; Linnaeus University, Sweden}\\
        Email: \{topfer, plasil, bures, hnetynka, krulis\}@d3s.mff.cuni.cz; danny.weyns@kuleuven.be
    }%
}

\maketitle
\thispagestyle{plain}

\begin{abstract}
Online machine learning (ML) is often used in self-adaptive systems to strengthen the adaptation mechanism and improve the system utility.
Despite such benefits, applying online ML for self-adaptation can be challenging, and not many papers report its limitations.
Recently, we experimented with applying online ML for self-adaptation of a smart farming scenario and we had faced several unexpected difficulties -- traps -- that, to our knowledge, are not discussed enough in the community. 
In this paper, we report our experience with these traps.
Specifically, we discuss several traps that relate to the specification and online training of the ML-based estimators, their impact on self-adaptation, and the approach used to evaluate the estimators.
Our overview of these traps provides a list of lessons learned, which can serve as guidance for other researchers and practitioners when applying online ML for self-adaptation.
\end{abstract}

\begin{IEEEkeywords}
Machine Learning, Online Machine Learning, Self-Adaptation, Traps.
\end{IEEEkeywords}

\section{Introduction}
\label{sec:introduction}
\input{introduction.tex}

\section{Running example and ML-DEECo}
\label{sec:background}
\input{background.tex}

\section{Experiments and results}
\label{sec:core}
\input{core.tex}

\section{Traps}
\label{sec:traps}
\input{traps.tex}

\section{Related work}
\label{sec:related-work}
\input{relwork.tex}

\section{Conclusion}
\label{sec:conclusion}
\input{conclusion.tex}

\section*{Acknowledgment}

This work was partially supported by the EU project
ExtremeXP grant agreement 101093164, partially by the European Research Council (ERC) under the European Union’s Horizon 2020 research and innovation programme (grant agreement No 810115), partially by Charles University institutional funding SVV 260698/2023, and partially by the Charles University Grant Agency project 269723.

\bibliographystyle{IEEEtran}
\bibliography{paper}

\end{document}

%% file: introduction.tex
The use of machine learning (ML) is rising in most areas of computer science including self-adaptive (SA) systems~\cite{weyns_2020}. Many approaches integrate the ML in the adaptation loop to estimate future or currently unobservable values (e.g., to predict resource usage), to prune the space of possible adaptations, or to create a model of the environment \cite{gheibi_applying_2021, saputri_application_2020}.

Most of these works use supervised machine learning (71\% of the works reviewed by Gheibi et al. in 2021~\cite{gheibi_applying_2021}) which learns
a function that maps inputs to outputs based on a set of labeled training examples (pairs of inputs and corresponding desired outputs).
In the majority of the applications, online ML is used, i.e., training the model incrementally as new data becomes available.

In the literature, mostly success stories of ML applications for SA are published. Only a small portion of such papers also focus on the limitations of using ML (18\% of the works reviewed by \cite{gheibi_applying_2021}\footnote{The reported limitations are listed in Table 7 in \cite{gheibi_applying_2021}.}). 
Nevertheless, in our recent work, we have encountered several other limitations (which we call \textit{traps} in this paper) when using online supervised ML for SA. For instance, one of the challenging tasks is how to properly evaluate how beneficial the use of ML in a specific SA application actually is.

Even though there are several works that give general recommendations to avoid mistakes when using ML (e.g.,~\cite{lones_pitfalls_2021}) and also explore the possible common mistakes when applying ML methods to a specific field, such as genomics~\cite{Whalen2021}, 
to the best of our knowledge, there is no paper 
that systematically focuses on the possible difficulties when applying ML for SA.

The goal of this paper is to share our experience with the ``unexpected hurdles'' -- traps -- that we encountered when applying ML for SA.
From this perspective, we aim to address the following research question:
\begin{itemize}
    \item[] \textit{What are the specific traps of applying online supervised ML for SA (experience using ML-DEECo)?}
\end{itemize}

To experiment with online supervised learning in self-adaptive systems, we developed a machine-learning-enabled component model called ML-DEECo~\cite{topfer_ensemblebased_2022} that allows modeling systems with ML-based predictions of the future. Alongside the component model, we provide its implementation as a Python framework~\cite{topfer_mldeeco_2022} to facilitate prototyping and experimenting with the systems.

We have used the ML-DEECo framework to implement simulations of several self-adaptive systems including use-cases from smart farming~\cite{abdullah_introducing_2022} (one of them we use in this paper as the running example) and Industry 4.0~\cite{topfer_ensemblebased_2022}. 

While prototyping the systems and during their evaluation, we encountered several traps related to the application of online ML for SA. 
For instance, if the relationship between the error of predictions and the utility of a system is not measured, but, this relationship is actually weak, the system may perform well even without the ML-based predictions. If an error in the predictions in such a setting is reduced, the utility of the system does not necessarily have to be improved.
The aim of this paper is to analyze and discuss the traps we encountered in our work and share the experience we gathered when handling them.

The remainder of this paper is structured as follows. In Section~\ref{sec:background} we introduce the running example and basics of the ML-DEECo component model. Section~\ref{sec:core} summarizes our experiments and their results. Section~\ref{sec:traps} discusses the traps that we encountered and provides insights to avoid them. Section~\ref{sec:related-work} presents related work, and Section~\ref{sec:conclusion} concludes the paper.

%% file: background.tex
\subsection{Smart field protection system (SFPS)}

In this paper (as we also did in our recent works, such as \cite{abdullah_introducing_2022, topfer_machine_2022}), we 
use scenarios from a running example employed in our recently finished ECSEL JU project AFarCloud\footnote{\url{https://www.ecsel.eu/projects/afarcloud}}.

Coming from the area of smart farming, the example introduces a self-adaptive system using autonomously flying drones for the protection of crops on a field against attacking birds. In this paper, we will refer to this system as SFPS (Smart Field Protection System). In essence, the drones patrol the field and by their presence (and the noise of their propellers) scare the birds away; if the field is not protected adequately, the birds damage the crops to an extent. To coordinate the protection of a field, the drones are assigned a specific position above the field at which they hover to scare away the birds. Obviously, the number of drones needed for full protection of the field is proportional to its size and to the number of attacking birds 

Since drones are battery-powered, they have to be recharged periodically at a charger. Further, we 
assume that there is a single charger associated with the field and that the number of drones that can be charged simultaneously is limited.

The main adaptation aspect of SFPS is the need to balance the trade-off between the number of drones required for protecting the field and the charging opportunities of drones. A specific challenge is that a drone needs to arrive at the charger before its battery runs out, and if the number of active drones exceeds the number of charging slots, the drone potentially needs to postpone its moving to the charger until a charging slot is likely to be available at the time of its arrival there.

Apparently, the adaptation decisions articulated in SFPS adaptation rules ought to benefit from using ML-based predictions of the future system state development such as the future battery level of a drone, and the availability of a charging slot at a specific time. From this perspective, we 
assess the utility of SFPS (efficiency of its functionality) by employing the following two metrics: 
\begin{enumerate*}
  \item The extent of damage to the crops, and
  \item the number of drones failing to arrive at the charger in time (and thus running out of battery).
\end{enumerate*}

\subsection{ML-DEECo}

To experiment with SA systems and online ML, we have developed a machine-learning-enabled component model ML-DEECo~\cite{topfer_ensemblebased_2022}. Building on the DEECo ensemble-based component model~\cite{bures_language_2020}, ML-DEECo enhances it with abstractions for estimating the future states (using supervised ML). 

For modeling and simulations of the SFPS,  
we use the abstractions of ML-DEECo.\footnote{Technically, for the experiments presented in this paper, we used our Python implementation of ML-DEECo~\cite{topfer_mldeeco_2022}.} All entities in SFPS (drones, birds, \dots) are modeled as ML-DEECo components, the states of which are represented by the values of their data fields.

Based on the state of the components and the state of the environment, the components can be grouped into \textit{ensembles}. As a key entity supporting the self-adaptation of the system at hand, an ensemble is a dynamically formed group of the system's components serving for their cooperation. 
In this work, we focus only on the way online ML is employed in the adaptation rules in both components and ensembles (for more details on these two concepts, we refer the reader to our previous works~\cite{bures_language_2020, topfer_ensemblebased_2022}).
A key abstraction to support online ML in ML-DEECo adaptation rules is \textit{estimator}. 
An estimator provides a prediction of a future value that can help make decisions encoded in an adaptation rule (change of the state of a component, ensemble formation, etc.). In other words, the value of the estimator is computed at run time using an online-trained ML model and is thus able to provide predictions of a future system state.

For instance, an estimator can be used to predict the future battery level of a drone.
Written in a simple specification language, an example of such an estimator is shown in a fragment of the specification of the \lstinline{Drone} component in Listing~\ref{lst:drone}.
Here, the \lstinline{Drone} component has three ordinary data fields: \lstinline{position}, \lstinline{battery} (the current battery level), and \lstinline{mode} (the current activity). To predict the future battery level of the drone, the \lstinline{futureBattery} estimator is defined. As we work with supervised ML, inputs and true outputs are necessary for the training of its underlying ML model.\footnote{In Section~\ref{sec:core}, we provide a description of how the training occurs in experiments with SFPS. For technical details of the training process, we refer the reader to our previous works~\cite{abdullah_introducing_2022, topfer_machine_2022}.}
This estimator uses the current values of \lstinline{battery} and \lstinline{mode} as inputs and the \lstinline{battery} in the future (after 1 to 200 time steps) as the true output. The true value of the battery level can be observed in the future, so it can be used for training the ML model. On line~\ref{inlst:future-bat-use}, the estimated battery in the 50th future time step is used to make an adaptation decision in the component.

\begin{lstlisting}[escapechar=|,breaklines=true,style=specification,label=lst:drone,caption=A fragment of Drone component type specification.]
component type Drone
  field position: Position
  field battery: ChargeLevel
  field mode: Enum { IDLE, TERMINATED, MOVING_TO_CHARGER, CHARGING,...}
  estimator futureBattery:
    input battery
    input mode
    output battery in T+<1,200>
  # other code ...
  # examples of estimator's employment:
  if futureBattery(50) < 0.3 then ...|\label{inlst:future-bat-use}|
\end{lstlisting}

%% file: core.tex
We focus here on two scenarios of SFPS functionality. Employing ML-DEECo concepts, we have done several experiments with them via a simulation.
In both scenarios, we use ML-based estimators to predict future values featuring in adaptation rules. Specifically, we use neural networks~(NNs)~\cite{Goodfellow-et-al-2016} as the ML model to provide the predictions.

\begin{figure*}
    \centering
    \includegraphics[width=\linewidth]{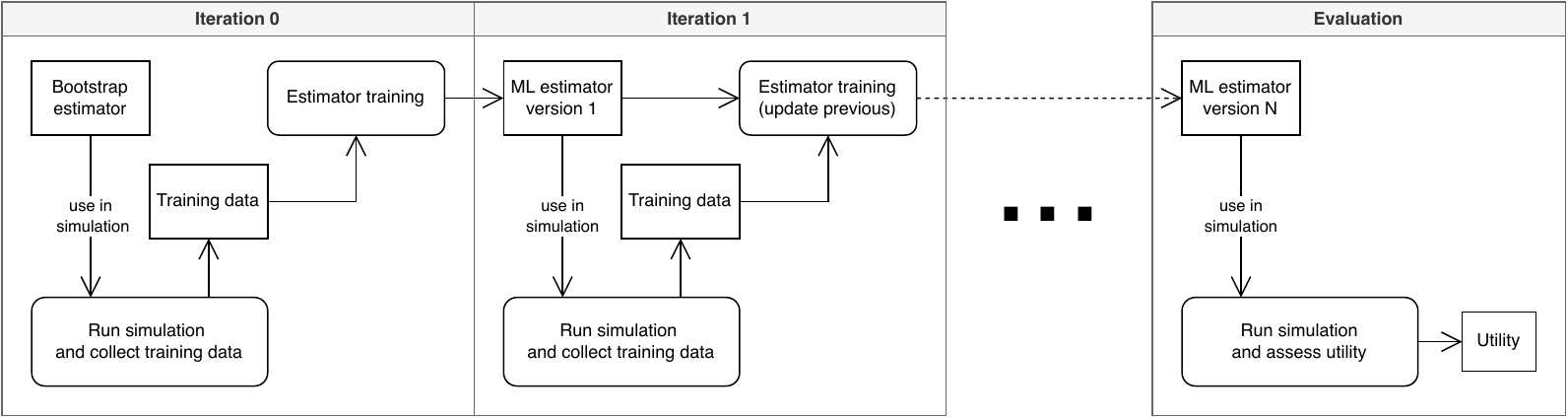}
    \caption{Activity diagram of simulating and estimator training in our experiments.}
    \label{fig:experiment-activity}
\end{figure*}

We train the estimators in iterations: Initially, we do not have a trained estimator, so we have to use a \textit{bootstrap estimator} instead  -- that can be realized by predicting a constant value\footnote{In our experience, the bootstrap constant can be chosen arbitrarily as long as it is adequate in the context of the system.}. First, we run the simulation with the bootstrap estimator for a certain period of time to collect relevant training data. Then, we use the data to train the first ML-based estimator. The simulation is then run with the newly trained estimator and more data is collected. The new data is then used to update the estimator (i.e., continue training the underlying ML model on the new data). This way, several iterations of simulating and training are performed. After that, either the last or the best-performing (explained below) version of the trained estimator is selected for evaluation. Figure~\ref{fig:experiment-activity} shows this process.

To evaluate the results, we need an assessment of how well SFPS is performing, measured by the \textit{utility} of SFPS. In the simulation, we use two metrics to measure the utility: \begin{enumerate*}
    \item The extent of damage to the field, and
    \item the number of drones that survived until the end of the simulation (their battery did not run out completely).
\end{enumerate*}
To assess whether ML-based estimators are helpful, we compare their effect with a baseline version of the scenario, where the ML-based estimator is replaced by a constant estimator (i.e., predicting a constant value).

Apart from the utility of the whole SFPS, we also assessed each of the estimators individually using standard ML evaluation techniques and metrics (such as accuracy and mean squared error). This was helpful in determining whether the estimator was able to predict the relevant values reasonably well. If the estimator did not perform well enough, we tried different ML algorithms and sets of hyperparameters to find a good estimator for the given task. As is common in ML, the data collected during the simulations were randomly split into training and testing data for evaluation of the estimator.

\subsection{Drone charging}
\label{sec:core:charging}

In the first scenario, we focus on charging of drones. Assuming that the number of available charging slots is less than the number of active drones, we coordinate the charging of drones using an adaptation rule. This rule is based on determining the expected battery level at the time when the drone in question can start charging. If the level is below a safety threshold, the drone is added to the FIFO waiting queue associated with the charger (nevertheless, the drone continues to protect the field). The drone starts moving to the charger (and is taken from the queue) with respect to the moment when a charging slot is expected to be available at the time of its arrival at the charger.
To evaluate the expected battery level in the adaptation rule, two estimators are applied: \lstinline{waitingTimeEstimator} and \lstinline{futureBatteryEstimator}. Whilst \lstinline{waitingTimeEstimator} predicts how long a particular drone will be listed in the waiting queue (before a free slot is available), \lstinline{futureBatteryEstimator} predicts the battery level after a given time. Furthermore, we also include the time needed to fly to the charger (linearly dependent on the distance to the charger). Since the position of the drone does not change when it protects the field (it is hovering at its dedicated position above the field), the time needed to fly to the charger does not change either. Thus, the key part of the rule is formed by the following predicate indicating that a drone requires charging:

\begin{lstlisting}[escapechar=|,breaklines=true,style=adaptation,label=lst:drone-charging,caption=Adaptation rule for drone charging.]
timeTillChargingStarts = waitingTimeEstimator(drone) + timeToFlyToCharger|\label{inlst:time-till-charging-starts}|

if futureBatteryEstimator(timeTillChargingStarts) < safetyThreshold then
        drone to be added to the queue
\end{lstlisting}

While designing \lstinline{futureBatteryEstimator} is relatively easy -- in our simulation, the battery consumption is constant, so the estimator can also be formulated without applying ML as a simple linear function, \lstinline{waitingTimeEstimator} is more complex since the waiting time depends on the behavior of the other drones and potentially other factors such as how far the drone is from the charger.

To evaluate the usefulness of ML-based estimators, we compared three concrete versions of \longlstinline{waitingTimeEstimator}: an ML-based estimator and two constant estimators predicting values $0$ (no waiting) and $100$ (a waiting time long enough to drain out the battery of a drone), respectively.
We evaluated the ML-based estimator and the two constant estimators using the two metrics of utility mentioned earlier in this section -- total damage to the fields, and the number of drones that survived until the end of the simulation.

\begin{figure*}
    \centering
    \includegraphics[width=0.88\linewidth]{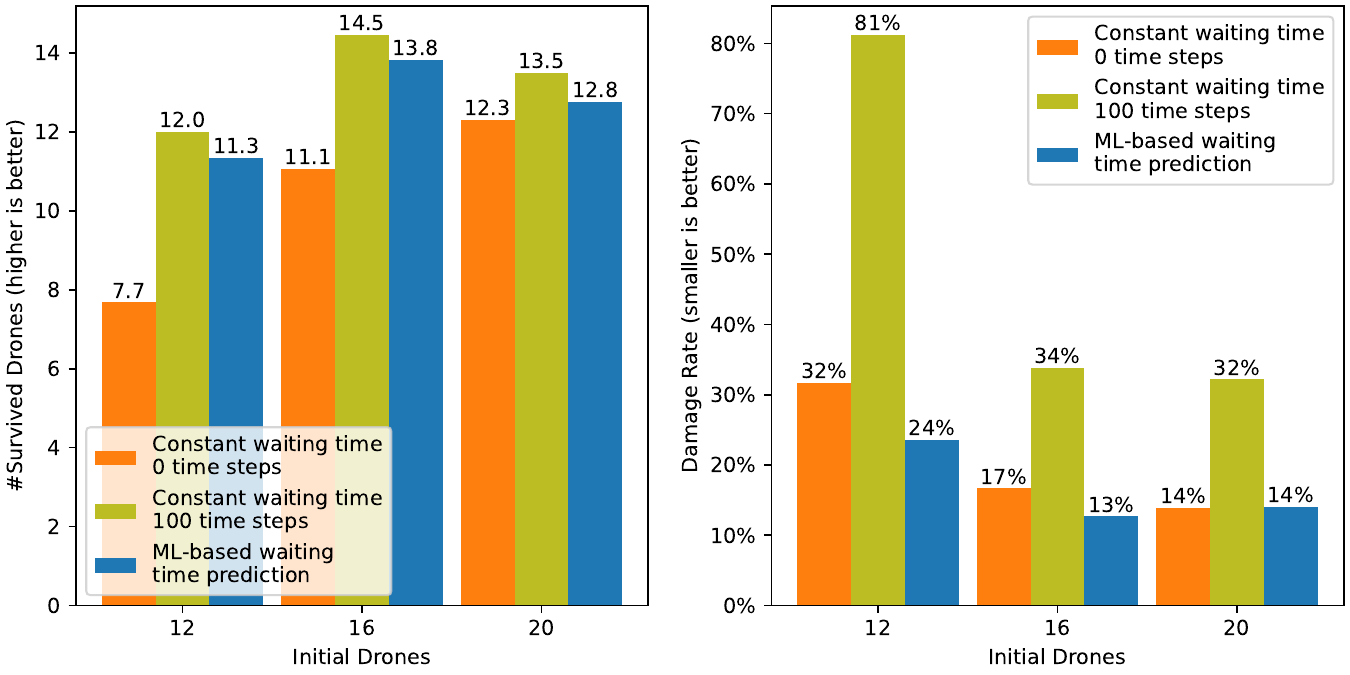}
    \caption{Comparison of survived drones and damage rate for constant and ML-based \footnotelstinline{waitingTimeEstimator} in Drone charging scenario.}
    \label{fig:drone_charging_summary}
\end{figure*}

The results of this experiment are summarized in Figure~\ref{fig:drone_charging_summary}, for details we refer to \cite{abdullah_introducing_2022, topfer_machine_2022}. When waiting time is set to constant 0 time steps, drones spend more time protecting the fields and less time charging. However, this also leads to overfilling of the queue associated with the charger, and, as a consequence, more drones run out of battery. When the waiting time is set to constant 100, drones are registered in the queue earlier, spending more time charging and less time protecting, which results in higher damage to crops in the field. When an ML-based estimator is employed, the results are comparable with the better of the baselines in both metrics of utility. Even though there is a slight decrease in the number of survived drones, the field damage rate improves notably.

\begin{figure}
    \centering
    \includegraphics[width=\columnwidth]{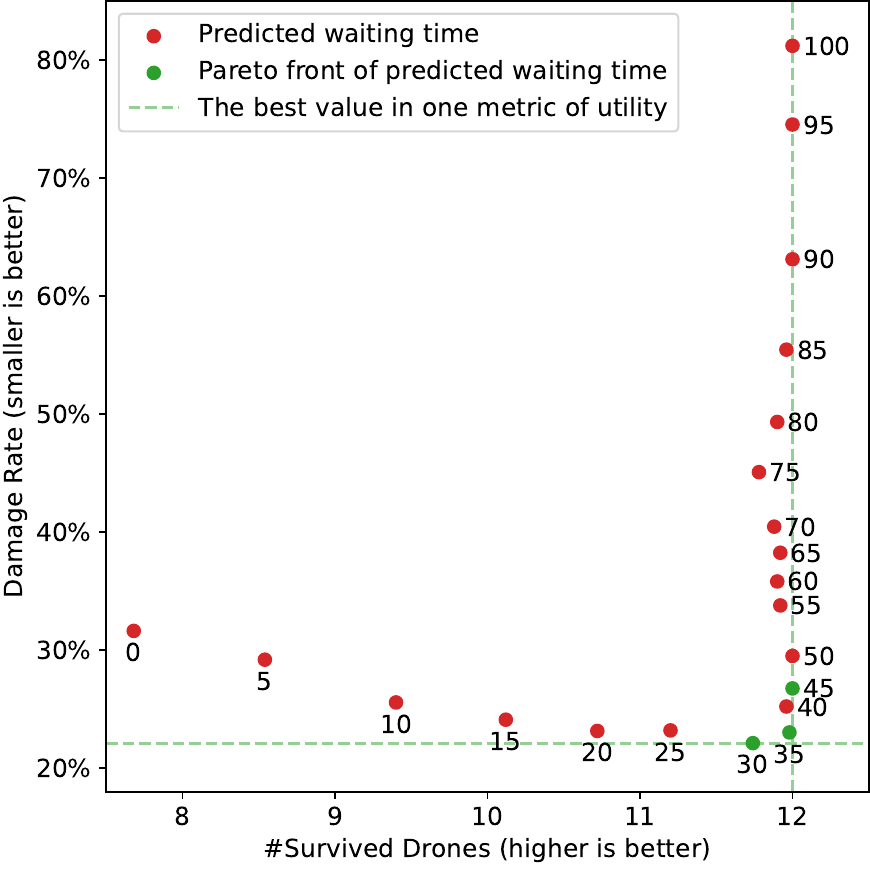}
    \caption{Comparison of survived drones and damage rate for different versions of the constant \footnotelstinline{waitingTimeEstimator} in Drone charging scenario. The data come from simulations with 12 drones. The Pareto-efficient~\cite{wiki-pareto-front} estimator versions are highlighted in green.}
    \label{fig:drone_charging_baselines}
\end{figure}

By further experimenting with the scenario, we did a more exhaustive grid search among the values for the parameter of the constant estimator. Figure~\ref{fig:drone_charging_baselines} shows the values of the metrics of utility for different values of the waiting time predicted as a constant. We found that constant estimators that predict values around $35$ 
performed not only better than the previously used constant estimators (with values $0$ and $100$) but also comparably to the ML-based estimator. In other words, always predicting that the waiting time will be 35 time steps gives similar results in terms of utility (damage rate and the number of survived drones) of SFPS as the ML-based estimator.

We came across similar results in several versions of the simulation scenario (different number of drones, slightly altered protocols for field protection, etc.). We discuss reasons for these results in Section~\ref{sec:traps}. 

\subsection{Field protection}
\label{sec:core:protection}

In the second scenario, we focus on adaptation that uses a prediction of the number of attacking birds in the future for balancing the number of active drones in field protection and the number of charging drones, that can protect the field later.
The key idea of adaptation logic is that when the predicted number of attacking birds in the near future is higher than the current number of attacking birds, it is worth partially sacrificing the protection drone activities in favor of their charging so they can better protect the field later when more attacking birds appear.
Here, it is assumed that the total number of drones is limited but sufficient enough to fully protect the field when no drone is currently charging. To lower the damage to the crops as much as possible, we want to charge the drones when the birds are not active and protect the field when the birds attack.

\begin{figure}
    \centering
    \includegraphics[width=\columnwidth]{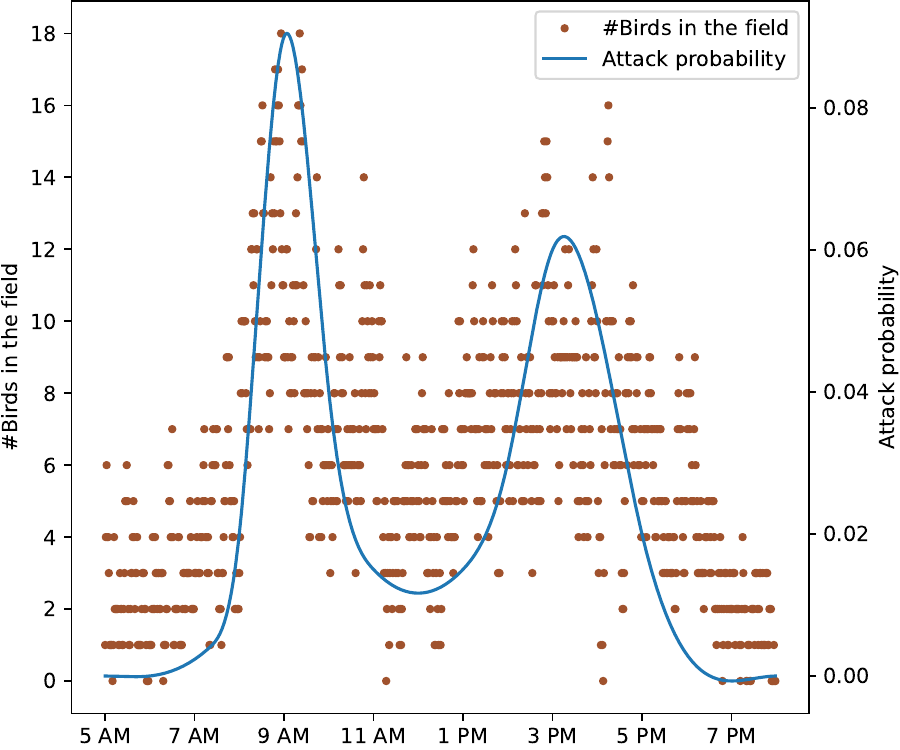}
    \caption{Attack probability of the birds and the number of birds detected in the field (per each minute) in a simulation of Field protection scenario with 100 birds.}
    \label{fig:attack-probability}
\end{figure}

Thus, via an estimator (\lstinline{futureBirdsEstimator}), the adaptation rule should benefit from predicting the number of birds attacking the field.
In this scenario, we assume that the bird activity varies throughout the day and its pattern repeats every day and that sensors in the fields detect the presence of birds and collect corresponding data. The true attack probability is modeled using a smooth function with two peaks (a bigger one around 9~AM, and a smaller one around 3~PM) and the attacks of birds are simulated according to this probability. The number of birds detected in the fields (i.e. the subject to the prediction by \lstinline{futureBirdsEstimator}) is thus highly correlated with the underlying attack probability. Both the attack probability and the number of detected birds in a simulated run are shown in Figure~\ref{fig:attack-probability}. Technically, \lstinline{futureBirdsEstimator} is constructed to use the current time of the day as input and predict the number of birds attacking the field in the near future\footnote{The \footnotelstinline{futureBirdsEstimator} predicts the number of birds attacking 150 minutes from now. The number was selected based on the time required to charge a drone -- the full charge of the battery takes 200 time steps, but since the drones do not run out of battery completely, the charging is usually faster.}.

The adaptation rule scales the number of protecting drones linearly with the number of currently attacking birds and the predicted number of attacking birds in the near future (the prediction from \lstinline{futureBirdsEstimator}). To achieve this technically, a safety threshold is computed -- with parameters $b$ (the intercept), $c$ (the multiplicative coefficient for current attacking birds), and $f$ (multiplicative coefficient for predicted attacking birds in the future) -- and a predicate is formed to decide whether a drone should be charged or should protect the field, as follows:

\begin{lstlisting}[escapechar=|,breaklines=true,style=adaptation,label=lst:field-protection,caption=Adaptation rule for field protection.]
safetyThreshold = |$b$| + |$c$| * currentBirds + |$f$| * futureBirdsEstimator()|\label{inlst:safety-threshold}|

if currentBattery - energyToFlyToCharger < safetyThreshold then
        drone to be added to the queue
else
        drone protects the field
\end{lstlisting}

We compared the above approach with
a baseline that scales the number of protecting drones linearly with the number of attacking birds. This is realized by setting $f=0$ in the formula above. Without the knowledge of the future, we use as many drones as possible to protect against the current peak of the number of attacking birds.

\begin{figure*}
    \centering
    \includegraphics[width=0.95\linewidth]{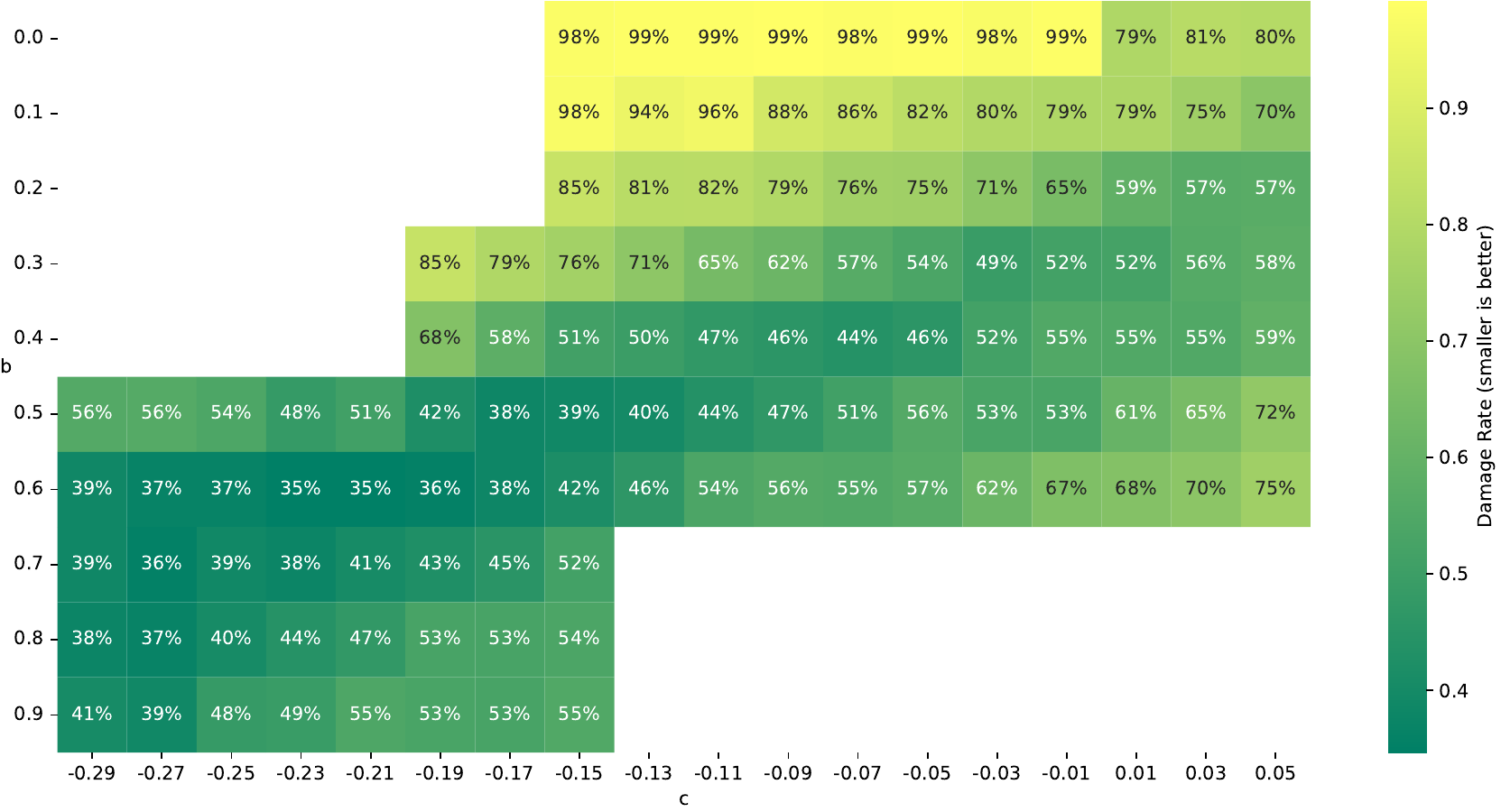}
    \caption{Results of the grid search for parameters $b$ and $c$ for the baseline approach in Field protection scenario.}
    \label{fig:grid}
\end{figure*}

We used a grid search to search for the coefficients ($b$, $c$, and $f$ in the formula) of both the ML-based and baseline approaches. 
We used the rate of damage dealt by the birds to the field as the utility for comparing the approaches. The results of the grid search for the best parameters for the baseline approach are shown in Figure~\ref{fig:grid}. The best baseline leads to the damage rate of 34.7\%. Several sets of parameters for the approach with the ML-based estimator of the future (\lstinline{futureBirdsEstimator}) resulted in smaller damage than the baseline, the best result was 25.6\%.

%% file: traps.tex
While implementing the simulations and running and 
experimenting with online ML for SA, and SFPS in particular, we encountered several traps that relate to the online ML for SA. In this section, we discuss them in more detail.

\subsection{Evaluation approach}

A remarkable trap that we encountered while experimenting with online ML for SA was the approach used for the evaluation.
As we designed the presented SA systems, there was usually no natural baseline to compare the usefulness of online ML-based estimators to (as opposed to introducing online ML to improve an existing SA system, in which case the natural baseline is the original version of the system).
In general, in the former case, constructing a baseline may introduce the problem that it might perform poorly and thus not be adequate enough for comparison with the adaptation featuring online ML-based estimators. 

This is exactly what happened in the Drone charging scenario (\ref{sec:core:charging}) in SFPS. We designed two baseline variants of \lstinline{waitingTimeEstimator} that we considered reasonable after a set of experiments -- a constant estimator predicting $0$ (no waiting) and, alternatively, a constant estimator predicting $100$ (long waiting time). The motivation behind these two baselines was that we aimed at optimizing the two metrics of utility (degree of damage of crops and the number of survived drones) and each baseline seemed to optimize its respective metric. The baseline constant $0$ was chosen to minimize the crop damage, and the baseline constant $100$ to maximize the number of survived drones.

However, when we further analyzed SFPS by performing a grid search of the possible constant variants of \lstinline{waitingTimeEstimator} (different waiting times), we found that even this simple constant estimator can perform well in terms of the utility of SFPS. By performing the grid search, we aimed to mimic a hypothetical system that would be explored and optimized by its creators.

Technically, the problem was that the two original baseline variants of \lstinline{waitingTimeEstimator} -- $0$ and $100$ -- were not located on the Pareto front~\cite{wiki-pareto-front}~(see Figure~\ref{fig:drone_charging_baselines}), though we implicitly expected them to be so.

Generally, implicitly assuming that the chosen baseline should perform well is a trap. As a way to avoid this trap, at least to a certain extent, we propose to design the baseline parametrically and perform a grid search (or some other optimization approach) to find the best parameters.

\subsection{Weak relationship between the error of estimator prediction and the utility of system}\label{sec:traps:weak-connection}

The utility of SFPS depends on many consecutive adaptation decisions as the adaptation mechanism balances the charging of drones and field protection throughout the whole run of a simulation. Furthermore, the ML-based estimators are only a part of the adaptation mechanism -- the adaptation rules also depend on the state of the components and the state of the environment. The error of an ML-based estimator (quality of its prediction) is thus only one of the factors influencing the adaptation efficiency; so having a good prediction may not necessarily improve the utility of SFPS. Overall, if the relationship (correlation) between the error of the predictions and the utility of a system is not significant, it might not be worth using ML-based estimators.

We fell into this trap in the Drone charging scenario of SFPS (\ref{sec:core:charging}). In our view, the complexity of SFPS and the weak relationship between the error of \lstinline{waitingTimeEstimator} and the number of survived drones (one of the SFPS utility metrics) is a reason why a constant estimator performed similarly to the ML-based estimator.

In other words, the usage of an ML-based estimator has to be leveraged. For instance, if it can provide insight regarding the future of a system state and the insight has a non-trivial impact on the overall utility of the system, then such an ML-based estimator can be useful. Otherwise, replacing the ML-based prediction with a simpler mechanism (such as using a constant value) can also lead to good results (see also Section~\ref{sec:traps:no-ml}).

\subsection{It is hard or impossible to design an estimator}

Sometimes, designing an estimator (defining its inputs and outputs, collecting relevant training data, etc.) is more complicated than anticipated, which also happened when experimenting with SFPS. 
Reasons for this include: The distribution of the predicted quantity is modified when the system adapts, the predicted quantity depends on some unknown (possibly immeasurable) variable(s) of the system, and wrong (irrelevant) data is used for the training of the estimator. These reasons are discussed in more detail below.

\subsubsection{The distribution of the predicted quantity is modified when the system adapts}

A key difference between \lstinline{waitingTimeEstimator} and \lstinline{futureBatteryEstimator} is that the distribution of the waiting time is modified when the system adapts while the distribution of future battery remains more or less the same. The waiting time in SFPS is heavily influenced by the adaptation decisions, which depend on the predictions of the waiting time by \lstinline{waitingTimeEstimator}. Thus, when \lstinline{waitingTimeEstimator} is trained on the data from one iteration (recall Figure~\ref{fig:experiment-activity}) and then employed in the next iteration, the predicted waiting time influences the adaptation decisions, causing a change in the actual waiting time. Thus, as a consequence, the predictions of \lstinline{waitingTimeEstimator} do not correspond to reality. On the other hand, the adaptation decisions do not influence the battery consumption of the drones so the predictions of \lstinline{futureBatteryEstimator} remain adequate.

Overall, it may be hard (or even impossible) to design an estimator with a low error for a quantity $Q$, assuming that the distribution of $Q$ is influenced through adaptation.
Using standard supervised ML algorithms (such as stochastic gradient descent for NNs)
is not enough in such a situation since the estimator is only able to learn the distribution from the training data, and it is not able to predict how the distribution is modified when the system adapts. Special techniques such as lifelong self-adaptation~\cite{gheibi_lifelong_2022} can be employed to deal with the drift in the distribution of quantity $Q$.

\subsubsection{Predicted quantity depends on an unknown (possibly immeasurable) variable}

In some cases, it is not possible to design a good enough estimator for the desired quantity. The reason might be that the predicted quantity does not really depend on the inputs used by the estimator.
There might be other variables that influence the quantity that we want to predict. 

If the variables are identified, they might be used as inputs of the model and the estimator might start performing adequately.
However, these variables might not be observable and in such cases, they cannot be used to design the estimator.

The trap here is that the creators of the system might not be aware of the hidden variables and thus they can be surprised that they are not able to design a good estimator.

For example, in SFPS, the true waiting time depends on the behavior of the other drones in the system. However, their behavior in the future is not observable at the time of the prediction. It is thus hard or even impossible to create good predictions based only on the observable variables (quantities) in the system.

This trap is related to the problem of confounding factors, which is elaborated in \cite{Whalen2021} and summarized in Section~\ref{sec:related-work:ml-traps}.

To mitigate this trap, all quantities that influence the prediction need to be identified. The variable selection process~\cite{wiki-feature-selection} can be used for selecting those that are relevant for the prediction.

\subsubsection{Adequacy of the training data} 

When designing an ML-based estimator, the adequacy of the data used for the training of the estimator must also be considered. In SA systems, the training data are often collected from the running system and later used to train the estimator. In such cases, extra attention must be paid to whether the collected data is relevant to the predicted quantity since using irrelevant data might skew the predictions and actually predict a slightly different quantity than originally wanted. This can obviously impact the utility of the system.

\begin{figure*}
    \centering
    \begin{subfigure}[t]{.47\textwidth}
        \centering
        \includegraphics[width=\linewidth]{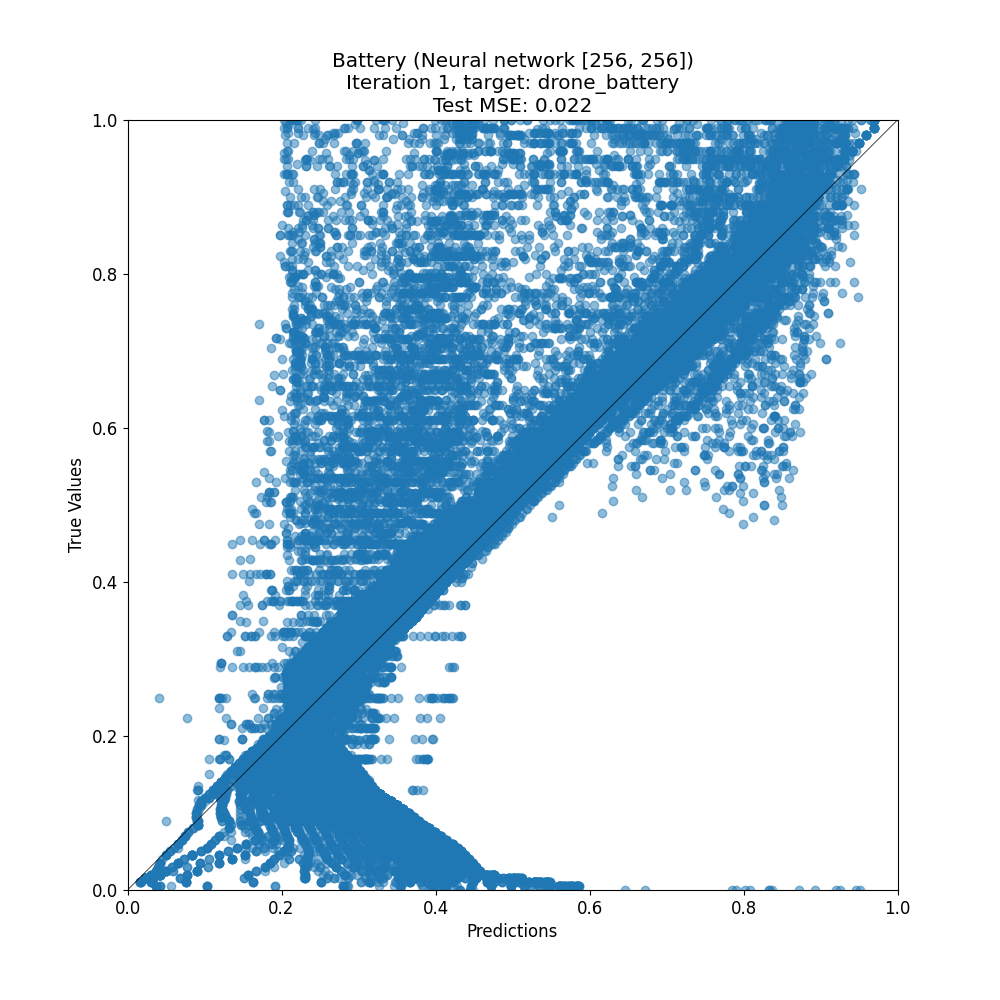}
        \caption{Estimator formed using inadequate data for training (including irrelevant data). The estimator does not perform well in this case.}
        \label{fig:traps:guard-bad}
    \end{subfigure}%
    \hfil%
    \begin{subfigure}[t]{.47\textwidth}
        \centering
        \includegraphics[width=\linewidth]{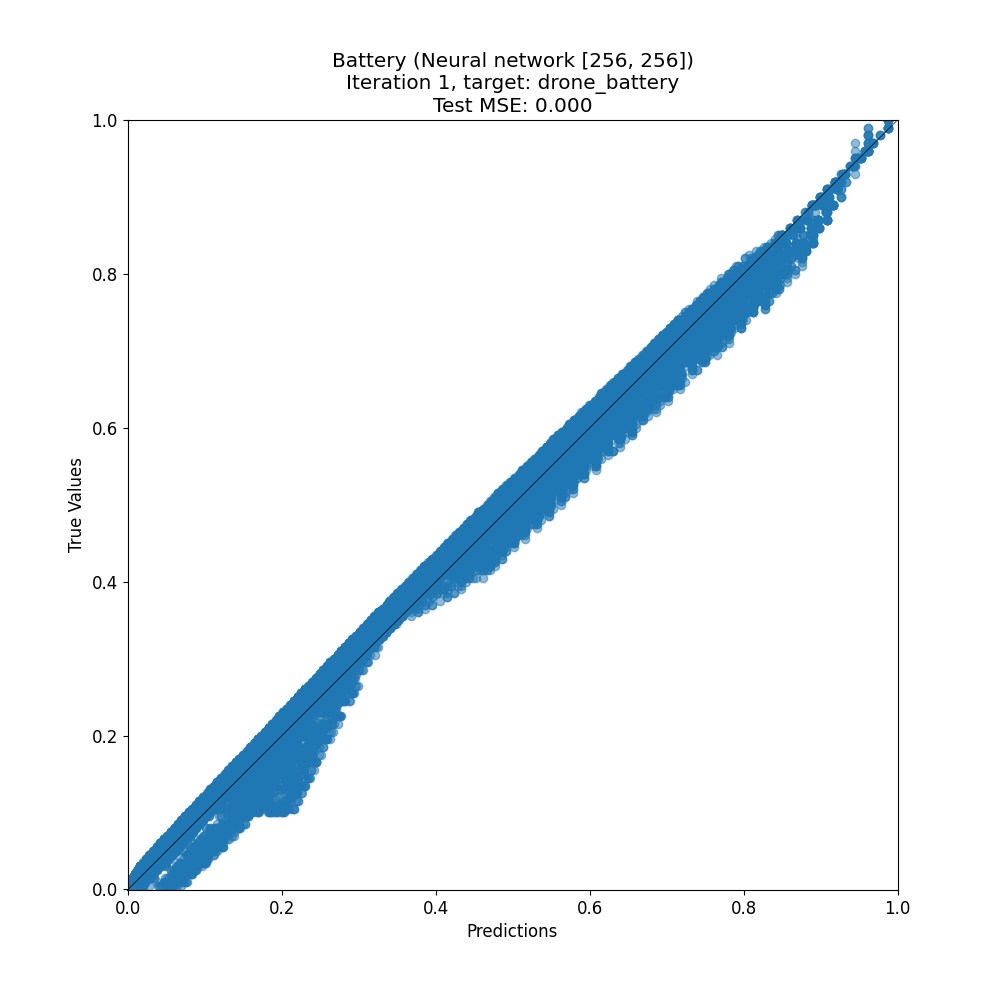}
        \caption{Estimator formed using adequate data for training -- the drone is not charging between collecting the inputs and the true outputs. The error of the estimator is much smaller.}
        \label{fig:traps:guard-good}
    \end{subfigure}
    \caption{Comparison of evaluation of two \footnotelstinline{futureBatteryEstimator} variants in SFPS. The horizontal axis show predicted values; the vertical axis the true values.}
    \label{fig:traps:guard}
\end{figure*}

We encountered this trap while working on \longlstinline{futureBatteryEstimator} 
that aims at predicting the battery level at the moment when a drone reaches the charger.
To that end, we collected the pairs of training data -- the current battery level as the input, and the battery level later in time as true output. The mentioned delay is determined by the expected prediction time horizon. However, \lstinline{futureBatteryEstimator} did not perform well, which we realized by examining the scatter plot of the predicted and true values, see Figure~\ref{fig:traps:guard-bad}, where all the points should ideally be near the diagonal.

When analyzing what the problem was, we found out that we incorrectly assumed that the drone was not charged in the period between observing the inputs and true outputs. When we used data from all the active drones (both those being charged in this period and those not being charged) for the training of \lstinline{futureBatteryEstimator}, the predicted quantity was ``battery level after a specific time.'' However, the quantity that we wanted to predict in the adaptation decision was ``battery level after a specific time, assuming the drone is not charged during that time''. To solve this problem, we only used the data from drones which were not charged in the time between observing inputs and true outputs for the training. When trained on adequate data, \lstinline{futureBatteryEstimator} performed much better as shown in Figure~\ref{fig:traps:guard-good}.

In summary, if inadequate data are used for the training of an estimator and incorrect assumptions are made regarding the training data, the estimator is likely to predict a wrong or different quantity than desired (and, consequently, affecting the adaptation decisions and utility of the system).

\subsection{Predicting quantities that are too complex or too random}\label{sec:traps:no-ml}

Another trap we encountered is using ML-based estimators for tasks that can actually be better performed by other approaches. When estimators are constructed for predicting quantities that are too complex or too random, an ML-based estimator may not be appropriate since a trivial random decision might not perform significantly worse. On the other hand, for tasks that are trivial or domain-specific, a domain expert may design a non-ML-based estimator that may often perform significantly better than an ML-based estimator.

\subsubsection{Constant or randomized estimator}

A constant or randomized estimator can give surprisingly good results. While experimenting with SFPS, we observed this when forming the adaptation rule in Listing~\ref{lst:drone-charging}. We used two approaches to compute \lstinline{timeTillChargingStarts} (line~\ref{inlst:time-till-charging-starts} in Listing~\ref{lst:drone-charging}), differing in the design of \lstinline{waitingTimeEstimator} -- one used a constant estimator, the other an ML-based estimator. By trying different constants for the former (for example those provided by grid search, Figure~\ref{fig:drone_charging_baselines}) we found a version of the constant \lstinline{waitingTimeEstimator} that saved all the drones in an SFPS run. In this particular case, the constant estimator was so good that exploiting ML in the design of an estimator was not useful. 

To summarize, in our experience, a constant or randomized estimator can be used in cases where there is a weak relationship between the error of the estimator and the utility of the system (see also Sec.~\ref{sec:traps:weak-connection}) and give surprisingly good results.

\subsubsection{Domain expert}

A way of designing a potentially good estimator is exploiting the knowledge of a domain expert to design the estimator manually.

In our work, we encountered this when working with \lstinline{futureBatteryEstimator} in the Drone charging scenario (\ref{sec:core:charging}).
In SFPS, drones consume energy from their batteries when moving and hovering in the air. In both these modes of operation, the energy consumption is constant (\longlstinline{movingBatteryConsumption} and \longlstinline{hoveringBatteryConsumption} respectively), but \longlstinline{hoveringBatteryConsumption} is lower than \longlstinline{movingBatteryConsumption}.

By exploiting this domain knowledge, two estimators can be designed -- an upper and lower bound -- for predicting the energy level of the battery at the time \lstinline{targetTime}:
\begin{lstlisting}[escapechar=|,breaklines=true,style=adaptation,label=lst:domain-expert,caption=Upper and lower bound \footnotelstinline{futureBatteryEstimator} designed\\ by a domain expert.]
estimator futureBatteryUpperBound(targetTime):
    timeDelta = targetTime - currentTime
    return currentBattery - timeDelta * hoveringBatteryConsumption

estimator futureBatteryLowerBound(targetTime):
    timeDelta = targetTime - currentTime
    return currentBattery - timeDelta * movingBatteryConsumption
\end{lstlisting}

In SFPS, the lower bound performed better. This is expected since the amount of energy consumed for flying to the charger is higher than the energy consumed for hovering above the field. So,  \lstinline{futureBatteryUpperBound} underestimates the energy needed to get to the charger and the drone can run out of battery. Clearly, the recommendations of domain experts should be considered with care. 

We also trained an ML-based estimator for performing the task of \lstinline{futureBatteryEstimator}. In this case, the error of the prediction was similar to \lstinline{futureBatteryLowerBound}, so we could have created the estimator manually, saving the time of training it. On the contrary, if \lstinline{futureBatteryUpperBound} was used instead of the ML-based estimator, the error of the predictions was higher.

A key benefit of manually designing an estimator is that it is usually better explainable~\cite{wiki-explainable,Camilli2022} than an ML-based one. Furthermore, a manually designed estimator can also be less computationally demanding than an ML-based estimator. Nevertheless, relying on domain expertise does not necessarily guarantee better results as illustrated with \lstinline{futureBatteryUpperBound}.

\subsection{Online ML affects training data}

Another trap comes from the online training of ML-based estimators. Such an estimator is retrained in iterations during the run of SFPS, see Figure~\ref{fig:experiment-activity}. 
Since the predictions provided by the estimator 
will influence the adaptation of the system,  
these predictions 
may also influence the data that will be used later for the training of the estimator. This can have a negative effect on the error of the predictions made by the estimator.

For example, in the Drone charging scenario of SFPS, \lstinline{waitingTimeEstimator} can oscillate between predicting low and high values. This corresponds to the behavior of SFPS when 
constant waiting times $0$ and $100$
are used, as summarized in Figure~\ref{fig:drone_charging_summary}. If low values are predicted, drones spend less time charging and more time protecting the field. The waiting time of a drone is longer in such a case since the other drones arrived at the charger with lower battery levels and needed more time to charge. Consequently, the estimator learns a longer waiting time and in the next iteration, drones arrive for charging much earlier -- needing less time to charge which also leads to shorter waiting times. The estimator is then trained on the shorter waiting times and SFPS oscillates between these two variants of waiting time prediction. In our experience, using training data from more than just the last iteration prevents this behavior.

A reason for this trap may be the  implementation of the ML-based estimators in SFPS -- we used here NNs. When retrained on new data, NNs might forget the training data from the prior iterations (known as catastrophic forgetting~\cite{goodfellow2015empirical}). Remembering data from the previous few iterations and using them together with the new data for the training usually helps.

\subsection{Relying on ML literature recommendations and not experimenting with the domain}\label{sec:traps:tuning}

The error of the estimator usually heavily depends on the ML algorithm used to make the predictions and the hyperparameters of the algorithm (the parameters that influence the ML algorithm itself, e.g., number of layers of a NN, learning rate, etc.). While one set of hyperparameters (e.g., based on recommendations in the literature such as~\cite{Goodfellow-et-al-2016}) can give reasonably good results, it is often worth experimenting with different sets of hyperparameters as they may provide even better results.

In the Field protection scenario (\ref{sec:core:protection}), we initially wanted to use NNs for the prediction of the activity of birds (\lstinline{futureBirdsEstimator}) as we had a good experience with them from the previous tasks. The learning task was seemingly easy, it was a regression with just a single input, so according to the Universal approximation theorem~\cite{HORNIK1991251}, it should be possible to model this function using a NN. However, we were not able to train the network to perform well using the standard stochastic gradient descent algorithm. Instead, we had to opt for a different ML algorithm
-- the $k$-nearest neighbors algorithm that turned out to work fine in this case.

Another example where we fell into this trap was the choice of the activation function of the last layer of an NN when designing \lstinline{waitingTimeEstimator} for the Drone charging scenario (\ref{sec:core:charging}). Initially, we started with an exponential activation function recommended by an expert colleague working on NNs. Later, by experimenting with different activation functions and hyperparameters of the NN, we found out that in our case, the softplus activation function performed better in terms of the overall utility of the system (but did not necessarily have a smaller error of the predictions).

Usually, it is good to try experiments with  several different ML algorithms, vary their hyperparameters, and observe the results. Again, the error of the estimator itself might not be the most important metric for the selection of the ML algorithm as the overall goal is to optimize the utility of the SA system.

%% file: relwork.tex
In this paper, we described potential traps when using online ML for SA based on our experiences. We divide related work into two categories -- papers focusing on potential traps when using ML in  general, and papers related to ML for SA that mention some limitations of using ML in this area.

\subsection{Traps/issues related to using ML}\label{sec:related-work:ml-traps}

There are several issues one can face when applying ML. Along these lines, Lones~\cite{lones_pitfalls_2021} provides a set of recommendations for researchers to avoid making mistakes when applying ML. These recommendations do not focus on any specific area but serve as valuable guidelines helpful to applying ML in general.

Whalen et al.~\cite{Whalen2021} focusing on the area of genomics performed a literature review, where they identified five pitfalls that often happen when applying ML in genomics. These authors state that ``the primary problem is that examples are assumed to be independent and identically distributed, but genomics is replete with violations of these assumptions.''
This impacts the ability of the ML model to generalize beyond the training data. In our experience, a similar problem also occurs in SA systems since the data collected from a running system are correlated (not independent), and the distribution of the data might also change over time. While there are works focused on the drift in the distribution of the data (such as~\cite{gheibi_lifelong_2022}), it is not clear how to deal with the dependent training data in SA systems; this is an interesting topic for future research.

Another pitfall reported by Whalen et al. is the problem of confounders. An ML model aims to predict outputs from a set of inputs, assuming that the outputs depend on the inputs. However, there might be another variable that is not observed (\textit{confounder}), but, it actually influences both the inputs and the outputs. In other words, since the inputs, as well as the outputs, may depend on the confounder, this induces a false association between the inputs and outputs. As Whalen et al. state ``this may have little or no effect on the accuracy of predictions, but it [can lead] to poor performance when the model is applied in a new context in which the confounder is absent or is distributed differently than in the original context.''

\subsection{Limitations of ML for SA}\label{sec:related-work:ml-sa}

The literature review by Gheibi et al.~\cite{gheibi_applying_2021} which focuses on applying ML for SA also compiles the limitations of ML reported by the reviewed papers (see Table 7 in~\cite{gheibi_applying_2021}). The most often reported limitations not included in our list of traps (hence complementing our list) are that the ML approach is not scalable and that ML requires a lot of computation time.

Another literature review by Saputri et al.~\cite{saputri_application_2020} states that ``there is very insufficient information on the limitations of ML for SA.'' These authors also mention the time required to train the ML model and the need to define the knowledge (the inputs and the outputs of the model) manually.

Several limitations and experiences are reported in Vaidhyanathan's thesis~\cite{vaidhyanathan_data-driven_2021} in sections titled ``Lessons learned''. Among other things, the author mentions the need for ML model selection and hyperparameter tuning as one of the limitations (explores as the trap in Section~\ref{sec:traps:tuning}) and the time required for learning (however, in his case, the overhead of ML was not dramatic and still enables getting ``Near-Real-Time Adaptations''). One of the limitations also mentioned is that ML can suffer from algorithmic bias (biased data leads to a biased learned model) with the suggestion to use quantitative verification to reduce the error rate of the ML model.

%% file: conclusion.tex
In this paper, we reported six traps (unexpected difficulties) that we encountered while performing research on applying ML in the design, analysis, and implementation of SA. The traps relate to both the design of ML-based estimators as well as their evaluation. We summarize our lessons learned with the six traps we encountered. 

The first and probably most important trap is \textit{the approach used for the evaluation}. To mitigate this trap we propose to design the baseline parametrically and perform a grid search or optimization approach to find the best parameters. 

The second trap is \textit{a weak relationship between the error of estimator prediction and the utility of system}. We advise investigating the usefulness of an ML-based estimator in terms of the utility of the system and considering alternative simpler mechanisms. 

The third trap is that \textit{designing an estimator may turn out to be more complicated than anticipated}. One issue here is the effect on the predictions due to changes in the distribution of data through adaptations or concept drift originating from the uncertainties at hand. This calls for the use of specific techniques such as lifelong self-adaptation~\cite{gheibi_lifelong_2022}. Another issue is that predictions may depend on an unknown (possibly immeasurable) variable or confounders. This calls for a careful variable selection process~\cite{wiki-feature-selection}. Finally, the training data collected from the running SA may not be adequate, resulting in inaccurate or even wrong predictions. Therefore, proper selection of training data is important. 

The fourth trap is that \textit{the predicted quantities are too complex or too random}. For such situations, an ML-based approach may not be appropriate. Instead, a simple standard non-ML-based estimator may be considered, and domain expertise may be exploited to create a solution.  

A fifth trap is that \textit{online training of ML-based estimators affects the training data}, which is common practice in SA. This may negatively affect the prediction errors. An important underlying cause may be the implementation of the estimator. Hence, taking care to avoid or minimize the effect of catastrophic forgetting~\cite{goodfellow2015empirical} is important.  

Finally, the sixth trap is \textit{relying on ML literature recommendations and not experimenting with the domain at hand}. Therefore, in order to make a well-informed decision about the solution to use, it is good practice to try experiments with several different ML algorithms, vary their hyperparameters, and observe the results. 

We hope that these insights will help researchers and engineers to create better ML-based solutions when realizing SA. In future work, we intend to explore the area of limitations and difficulties when working with ML for SA more systematically and in-depth. This could be realized in the form of a systematic literature review focused on the difficulties and issues in this area reported by others.